# Utilizing unsupervised learning to improve sward content prediction and herbage mass estimation

Albert P.[1,2], Saadeldin M.[2,3], Narayanan B.[2,3], Mac Namee B.[2,3], O'Connor N.[1,2], Hennessy D.[2,4], O'Connor A.H.[2,4], McGuinness K.[1,2]
[1]School of Electronic Engineering, Dublin City University, Whitehall, Dublin 9, Ireland; [2]VistaMilk Research Centre, Animal and Grassland Research and Innovation Centre, Moorepark, Fermoy, Co. Cork, Ireland; [3]School of Computer Science, University College Dublin, Belfield, Dublin 4, Ireland; [4]Teagasc, Animal and Grassland Research and Innovation Centre, Moorepark, Fermoy, Co. Cork, Ireland;

**Abstract**
Sward species composition estimation is a tedious one. Herbage must be collected in the field, manually separated into components, dried and weighed to estimate species composition. Deep learning approaches using neural networks have been used in previous work to propose faster and more cost efficient alternatives to this process by estimating the biomass information from a picture of an area of pasture alone. Deep learning approaches have, however, struggled to generalize to distant geographical locations and necessitated further data collection to retrain and perform optimally in different climates. In this work, we enhance the deep learning solution by reducing the need for ground-truthed (GT) images when training the neural network. We demonstrate how unsupervised contrastive learning can be used in the sword composition prediction problem and compare with the state-of-the-art on the publicly available GrassClover dataset collected in Denmark as well as a more recent dataset from Ireland where we tackle herbage mass and height estimation.

**Keywords**: biomass prediction, herbage mass prediction, unsupervised learning, clover

**Introduction**
Developing tools to help estimate biomass yield can improve decision making at the farm level. By fixing nitrogen (N) from the air into soils, white clover (*Lolium perenne* L.) has been shown (Nyfeler *et al*., 2009) to be an important element of pasture sward management. White clover is an important motivator in the cow's diet, increasing dry matter (DM) intake and milk production (Egan *et al*., 2018). A regular estimation of the expected herbage mass and biomass composition of the sward would provide a stepping stone towards targeted N fertilization, reducing the cost for the farmer and the environmental impacts of fertiliser (Ju *et al*., 2004). Estimation tools from canopy view images using neural networks have shown to be accurate but highly sensitive to different herbages with different visual characteristics (Narayanan *et al*., 2020). Reducing the amount of GT data required to train these models is essential to quickly adapt to different climates and geographic locations. Skovsen *et al*. (2019) proposed using synthetically generated images to train a segmentation network paired with a linear regression model for composition detection and Albert *et al*. (2021) automatically labelled un-GT images to augment the training dataset. In this paper, we use un-GT images to train an unsupervised model using contrastive learning (Phuc *et al*., 2020) which provides stronger initial weights to a neural network model performing composition detection. We show that this allows us to reduce biomass prediction error rates using un-GT images when compared to state-of-the-art algorithms. We compare with the state-of-the-art on the publicly available GrassClover dataset as well as an Irish dataset where we also look at biomass weight and herbage height estimation as well as predicting on images captured using phones.

**Materials and methods**

We aim to evaluate whether unsupervised learning can be used on un-GT images to bridge the gap between a partially GT dataset and a fully GT one. To do so, we use two image datasets to address biomass prediction and herbage weight estimation from canopy view images. The GrassClover image dataset gathered in Denmark in 2018 (Skovsen *et al.*, 2019) is composed of 152 fully GT biomass images (100 for training, 52 for validation) with the biomass composition comprising grass, weed, red clover and white clover. More images are available for test submission in an online challenge (https://competitions.codalab.org/competitions/21122), and 31,000 un-GT images are also provided. The second dataset was gathered in 2020 in Ireland (Hennessy *et al.*, 2021) using a Canon camera and is composed of 528 GT images with the biomass composition comprising grass, weed and clover. This dataset additionally provides ground-truth for the herbage mass (kg DM ha$^{-1}$) and herbage height post-cutting (cm). Additional views taken from a phone are also provided for evaluation purposes. We consider 52 images for training, 104 for validation and 372 for testing. A further 594 un-GT images are provided. For the unsupervised algorithm, we use the i-Mix algorithm (Lee *et al.*, 2021). This algorithm aims to compare two data augmented views of the same image against different images from the same mini-batch, promoting the learning of visual features useful to differentiate between images. After completing the unsupervised learning phase, we use the parameters (weights) learned to initialize the neural network before starting the supervised training phase. In the supervised training phase, we train the network to predict the species composition in a given dataset by minimizing a root mean square error (RMSE) objective over the training data. The network then outputs percentage values for weed, clover and grass and two values between 0 and 1 for the normalized herbage weight and height (Irish data only). All instructions to reproduce our results are available at https://git.io/JMrY1.

**Results and discussion**

We compare with state-of-the-art algorithms proposing low supervision alternatives to solve the biomass estimation problem. In Table 1, we report the RMSE of our approach and of Skovsen *et al.* (2019) (generating synthetic images), Narayanan *et al.* (2020) (using strong data augmentation) and Albert *et al.* (2021) (automatically labeling the unlabeled set) on the GrassClover dataset.

Table 1: Biomass prediction results on the GrassClover dataset, we bold the best results

|  | Grass | Any clover | White clover | Red clover | Weeds | Avg. |
| --- | --- | --- | --- | --- | --- | --- |
| Skovsen et al. | 9.05 | 9.91 | 9.51 | 6.68 | 6.50 | 8.33 |
| Narayanan et al. | 8.64 | 8.73 | 8.16 | 10.11 | 6.95 | 8.52 |
| Albert et al. | 8.78 | 8.35 | 7.72 | 7.35 | 7.17 | 7.87 |
| Ours | 8.02 | 7.31 | 7.74 | 8.22 | 6.61 | 7.54 |

For the Irish clover dataset (Hennessy *et al.*, 2021), we predict the herbage mass and multiply it by the predicted biomass percentage to evaluate the mass per species. We evaluate the Herbage RMSE (HRMSE) with regards to the ground-truth. We additionally report the Herbage Relative Error $HRE = \frac{pred}{gt}$ as in O'Donovan *et al.* (2002) where *gt* is the ground-truth value and *pred* is the predicted value by the network. Finally, HE is height prediction error (RMSE) in cm.

**Conclusion**

By initializing the neural network on unsupervised images, we reduce the prediction error when training with a limited amount of GT images. This opens the possibility of quickly adapting an herbage mass and composition algorithm to an unseen environment using a limited quantity of labor intensive GT images paired with a large quantity of un-GT data. Our results are comparable to Albert *et al*. (2021) without the need to generate synthetic images.

Table 2: Biomass prediction results on the Irish clover dataset. The top two rows indicate test results on the Canon images while the bottom two rows report results on held out phone images.

|  | HRMSE | | | | | HRE | RMSE | | | | HE |
|---|---|---|---|---|---|---|---|---|---|---|---|
|  | Total | Grass | Clover | Weeds | Avg. |  | Grass | Clover | Weeds | Avg. |  |
| Albert et al. | 230.10 | 220.84 | 34.86 | 27.13 | 94.28 | 1.14 | 4.81 | 4.75 | 3.42 | 4.33 | 2.15 |
| Ours | 229.12 | 218.02 | 37.65 | 29.21 | 94.96 | 1.09 | 4.58 | 4.22 | 3.44 | 4.08 | 2.03 |
| Albert et al. | 226.59 | 215.85 | 36.28 | 27.00 | 93.04 | 1.31 | 5.44 | 5.08 | 3.70 | 4.74 | 1.80 |
| Ours | 236.87 | 221.22 | 27.18 | 34.61 | 94.34 | 1.03 | 4.03 | 4.11 | 4.72 | 4.28 | 2.02 |

**Acknowledgements**

This publication has emanated from research conducted with the financial support of Science Foundation Ireland (SFI) under grant numbers SFI/15/SIRG/3283 and SFI/12/RC/2289P2.